%% file: main.tex
\documentclass[letterpaper, 10 pt, conference]{ieeeconf} 
\IEEEoverridecommandlockouts                             

\usepackage{amssymb}
\usepackage{amsmath}
\usepackage{commath}
\input{h2t_def}

\usepackage{booktabs,colortbl}
\usepackage{multirow}
\usepackage{graphicx}
\usepackage{fdsymbol}
\usepackage{tikz}
\usepackage{soul}
\usepackage{microtype}
\usepackage[scr=boondox]{mathalpha}
\let\labelindent\relax 
\usepackage[inline]{enumitem}
\usepackage{hyperref}
\enablefinalversion

\newcommand{\greymidrule}{\arrayrulecolor{black!30}\midrule[0.1pt]\arrayrulecolor{black}}

\newcommand{\Lukasiewicz}{\L{}ukasiewicz\xspace}
\newcommand{\lukweakconj}{\wedge_\text{\L{}}}
\newcommand{\relcondition}{\gamma}
\newcommand{\fuzzy}[1]{\tilde{#1}}

\newcommand{\textallenbefore}{\emph{before}}
\newcommand{\textallenafter}{\emph{after}}
\newcommand{\textallenmeets}{\emph{meets}}
\newcommand{\textallenmetby}{\emph{met\,by}}
\newcommand{\textallenoverlaps}{\emph{overlaps}}
\newcommand{\textallenoverlappedby}{\emph{overlapped\,by}}
\newcommand{\textallenstarts}{\emph{starts}}
\newcommand{\textallenstartedby}{\emph{started\,by}}
\newcommand{\textallenfinishes}{\emph{finishes}}
\newcommand{\textallenfinishedby}{\emph{finished\,by}}
\newcommand{\textallenduring}{\emph{during}}
\newcommand{\textallencontains}{\emph{contains}}
\newcommand{\textallenequals}{\emph{equals}}

\newcommand{\allenrel}{R^\text{\rotatebox{90}{\,I}}}
\newcommand{\allenbefore}{\allenrel_\text{\textallenbefore}}
\newcommand{\allenafter}{\allenrel_\text{\textallenafter}}
\newcommand{\allenmeets}{\allenrel_\text{\textallenmeets}}
\newcommand{\allenmetby}{\allenrel_\text{\textallenmetby}}
\newcommand{\allenoverlaps}{\allenrel_\text{\textallenoverlaps}}
\newcommand{\allenoverlappedby}{\allenrel_\text{\textallenoverlappedby}}
\newcommand{\allenstarts}{\allenrel_\text{\textallenstarts}}
\newcommand{\allenstartedby}{\allenrel_\text{\textallenstartedby}}
\newcommand{\allenfinishes}{\allenrel_\text{\textallenfinishes}}
\newcommand{\allenfinishedby}{\allenrel_\text{\textallenfinishedby}}
\newcommand{\allenduring}{\allenrel_\text{\textallenduring}}
\newcommand{\allencontains}{\allenrel_\text{\textallencontains}}
\newcommand{\allenequals}{\allenrel_\text{\textallenequals}}

\newcommand{\fallenrel}{\fuzzy{R}^\text{\rotatebox{90}{\,I}}}
\newcommand{\fallenbefore}{\fallenrel_\text{\textallenbefore}}

\newcommand{\fallenmeets}{\fallenrel_\text{\textallenmeets}}

\newcommand{\fallenoverlaps}{\fallenrel_\text{\textallenoverlaps}}

\newcommand{\fallenstarts}{\fallenrel_\text{\textallenstarts}}

\newcommand{\fallenfinishes}{\fallenrel_\text{\textallenfinishes}}

\newcommand{\fallenduring}{\fallenrel_\text{\textallenduring}}

\newcommand{\fallenequals}{\fallenrel_\text{\textallenequals}}
\newcommand{\pointrel}{R^\bullet}
\newcommand{\fpointrel}{\fuzzy{R}^\bullet}
\newcommand{\pointbefore}{\pointrel_\text{\textallenbefore}}
\newcommand{\pointafter}{\pointrel_\text{\textallenafter}}
\newcommand{\pointequals}{\pointrel_\text{\textallenequals}}
\newcommand{\fpointbefore}{\fpointrel_\text{\textallenbefore}}
\newcommand{\fpointafter}{\fpointrel_\text{\textallenafter}}
\newcommand{\fpointequals}{\fpointrel_\text{\textallenequals}}
\newcommand{\constrs}{\mathscr{c}}
\newcommand{\constrss}{c}
\newcommand{\Constrs}{\mathscr{C}}
\newcommand{\Constrss}{C}
\newcommand{\interval}{i}
\newcommand{\intervala}{\interval_1}
\newcommand{\intervalb}{\interval_2}
\newcommand{\intstart}[1]{#1^-}
\newcommand{\intend}[1]{#1^+}
\newcommand{\act}{\alpha}
\newcommand{\acta}{\act_1}
\newcommand{\actb}{\act_2}
\newcommand{\actas}{\intstart{\acta}}
\newcommand{\actbs}{\intstart{\actb}}
\newcommand{\actae}{\intend{\acta}}
\newcommand{\actbe}{\intend{\actb}}
\newcommand{\actobs}{a}

\newcommand{\smallno}{\epsilon}
\newcommand{\eqmargin}{m^{=}}
\newcommand{\gmm}{M}
\newcommand{\gmmdata}{T}
\newcommand{\gmmcomponentsymbol}{\kappa}
\newcommand{\gmmcomponentssymbol}{K}
\newcommand{\gmmcomponent}[2]{\gmmcomponentsymbol_{#1, #2}}
\newcommand{\gmmcomponents}[1]{\gmmcomponentssymbol_{#1}}

\newcommand{\gmmcomponentsne}[1]{\gmmcomponentssymbol^{\neq}_{#1}}
\newcommand{\gmmpdfsymbol}{f}
\newcommand{\gmmpdf}[1]{\gmmpdfsymbol_{#1}}
\newcommand{\gmmpdfne}[1]{\gmmpdfsymbol^{\neq}_{#1}}
\newcommand{\gmmdataasbs}{\gmmdata^{\actas}_{\actbs}}
\newcommand{\gmmdataasbe}{\gmmdata^{\actas}_{\actbe}}
\newcommand{\gmmdataaebs}{\gmmdata^{\actae}_{\actbs}}
\newcommand{\gmmdataaebe}{\gmmdata^{\actae}_{\actbe}}
\newcommand{\gmmasbs}{\gmm^{\actas}_{\actbs}}
\newcommand{\gmmasbe}{\gmm^{\actas}_{\actbe}}
\newcommand{\gmmaebs}{\gmm^{\actae}_{\actbs}}
\newcommand{\gmmaebe}{\gmm^{\actae}_{\actbe}}
\newcommand{\gmmsab}{\mathcal{\gmm}^{\acta}_{\actb}}

\expandafter\def\expandafter\normalsize\expandafter{%
    \normalsize%
    \setlength\abovedisplayskip{2mm}%
    \setlength\belowdisplayskip{2mm}%
    \setlength\abovedisplayshortskip{0mm}%
    \setlength\belowdisplayshortskip{1mm}%
}

\title{\LARGE \bf
Learning Symbolic and Subsymbolic Temporal Task \\ %
Constraints from Bimanual Human Demonstrations%
}

\author{Christian Dreher and Tamim Asfour%
\thanks{The research leading to these results has received funding from the German Research Foundation (DFG) within the SFB 1574 and the Carl Zeiss Foundation through the AgiProbot project.}
\thanks{The authors are with the Institute for Anthropomatics and Robotics, Karlsruhe Institute of Technology, Karlsruhe, Germany. {\tt \{c.dreher, asfour\}@kit.edu}}%
}

\begin{document}

\maketitle
\thispagestyle{plain}
\pagestyle{plain}
\thispagestyle{empty}
\pagestyle{empty}

\begin{abstract}
Learning task models of bimanual manipulation from human demonstration and their execution on a robot should take temporal constraints between actions into account. 
This includes constraints on (i) the symbolic level such as precedence relations or temporal overlap in the execution, and (ii) the subsymbolic level such as the duration of different actions, or their starting and end points in time. 
Such temporal constraints are crucial for temporal planning, reasoning, and the exact timing for the execution of bimanual actions on a bimanual robot. 
In our previous work, we addressed the learning of temporal task constraints on the symbolic level and demonstrated how a robot can leverage this knowledge to respond to failures during execution.
In this work, we propose a novel model-driven approach for the combined learning of symbolic and subsymbolic temporal task constraints from multiple bimanual human demonstrations. 
Our main contributions are a subsymbolic foundation of a temporal task model that describes temporal nexuses of actions in the task based on distributions of temporal differences between semantic action keypoints, as well as a method based on fuzzy logic to derive symbolic temporal task constraints from this representation.
This complements our previous work on learning comprehensive temporal task models by integrating symbolic and subsymbolic information based on a subsymbolic foundation, while still maintaining the symbolic expressiveness of our previous approach. 
We compare our proposed approach with our previous pure-symbolic approach and show that we can reproduce and even outperform it.
Additionally, we show how the subsymbolic temporal task constraints can synchronize otherwise unimanual movement primitives for bimanual behavior on a humanoid robot.
\end{abstract}

\section{Introduction}

\newcommand{\sttc}{STTC\xspace}
\newcommand{\ssttc}{SSTTC\xspace}
\newcommand{\sttcs}{STTCs\xspace}
\newcommand{\ssttcs}{SSTTCs\xspace}

Humans have the inherent ability to learn tasks by observing others.
Especially assistive humanoid robots that are developed to engage in interactions with humans should also have this ability.
This is not only due to the assumption that users will not have the necessary programming skills but also because teaching a task as humans teach each other is the most intuitive interface for robot programming. 
Robot programming by demonstration has been seen as a powerful mechanism for reducing the complexity of the search space for learning. 
It also provides an implicit means for teaching a robot new skills so that explicit and tedious programming by a human can be minimized or eliminated~\cite{billard2008robot}. 
An essential aspect of this is the question of extracting as much task-specific information as possible from a few human demonstrations. 
Observed temporal constraints between actions and temporal keypoints of actions in manipulation tasks are an important part of this information. 
The temporal constraints can be symbolic (\eg certain actions must occur before, after, during, etc. other actions), and subsymbolic (\eg the start of an action always has a certain time delay to the end of another action). 
These symbolic temporal task constraints (\sttcs) are crucial for action sequencing, temporal planning, or general temporal reasoning.
Subsymbolic temporal task constraints (\ssttcs) play an important role in the coordination and synchronization of actions or their underlying movement primitives (MPs), how long the execution of an action usually takes, and how much temporal flexibility is available during execution.

\begin{figure}[t]
    \vspace*{-3.5mm}
	\centering
	\includegraphics[width=\linewidth]{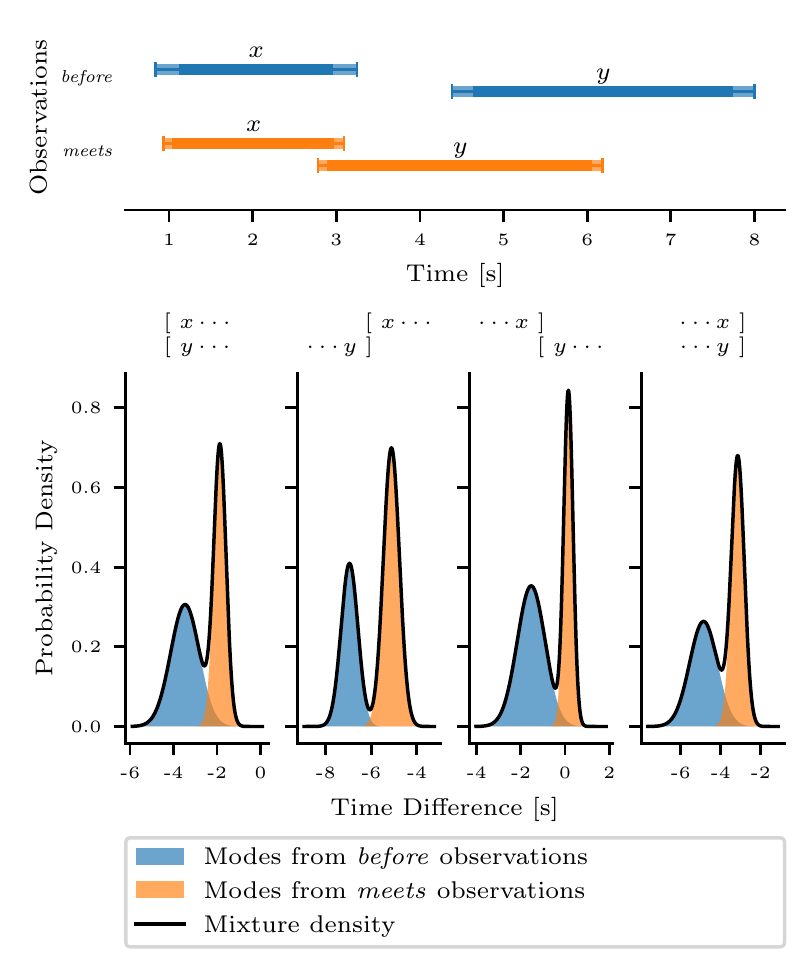}
    \vspace*{-7mm}
	\caption{Basic idea of our approach. Upper: Two modes of task execution where actions $x$ and $y$ either follow a \emph{before} or \emph{meets} pattern. 
    Lower: Instead of formulating temporal task constraints between two actions $x$ and $y$ semantically using Allen relations (\emph{before} or \emph{meets} in this case), we represent them as Gaussian mixture models of temporal differences between action's starts and ends observed in human demonstrations.
    This allows for a combined subsymbolic and symbolic representation.}\label{fig:facefig}
    \vspace*{-5mm}
\end{figure}

In our previous work \cite{dreher2022learning}, we presented an approach to obtain \sttcs.
Observed symbolic temporal relations between actions were collected in a graph while subsymbolic information was discarded.
In this work, we present an approach based on temporal differences of semantic action keypoints as a knowledge base instead of symbolic interval relations to account for learning and representing both \sttcs and \ssttcs.
We hypothesize that all desired temporal constraints mentioned before can be inferred by analyzing temporal differences between semantic action keypoints.
The proposed model can extract both
\begin{enumerate*}[label=(\roman*)]
    \item the temporal constraints on the symbolic level as in our previous work \cite{dreher2022learning} and 
    \item the subsymbolic information of such constraints.
\end{enumerate*}
Our contributions are
\begin{enumerate*}[label=(\roman*)]
    \item a temporal-differences-driven formulation of \ssttcs that models the temporal nexuses between semantic action keypoints in manipulation tasks,
    \item a novel approach to derive \sttcs from such a model of temporal differences, and
    \item an evaluation of the formulation of our temporal task model on two publicly available datasets for learning temporal constraints of complex bimanual manipulation tasks, one manually labeled, and one synthetic, as well as a showcase demonstrating the usefulness of \ssttcs.
\end{enumerate*}

\section{Related Work}\label{s:sota}

From a methodological perspective, related works concerned with modeling temporal constraints in general can be classified into three branches: Two explicit ones based on Allen's Interval Algebra or Linear Temporal Logic, and implicit temporal representations.
\emph{Allen's Interval Algebra} \cite{allen1983maintaining} is a well-established formalism with the core idea of exhaustively defining all possible relations two intervals can be in.
The other well-established formalism is \emph{Linear Temporal Logic} (LTL) \cite{pnueli1977temporal} which is very similar to first-order logic and was originally proposed for formal system verification.
A discussion on the relationship between both formalisms was carried out by Roşu and Bensalem \cite{rosu2006allen}.
The last branch is \emph{implicit models}, where temporal or spatio-temporal features are autonomously inferred and learned via a variety of machine learning models.
That being said, we discuss relevant works with a focus on the purpose rather than on the methodology.
First, we discuss other works concerned with learning temporal constraints in Section~\ref{ss:rw-temp-constraints}. 
Next, we position our work in the domain of planning under temporal constraints in Section~\ref{ss:rw-temp-planning} and conclude with other relevant works beyond these research areas in Section~\ref{ss:rw-causal-constraints}.

\subsection{Learning Temporal Constraints}\label{ss:rw-temp-constraints}

The earliest works concerned with learning temporal constraints from human demonstrations focused on precedence relations or action sequences only.
These can be seen as a subset of Allen's interval algebra where only \textallenbefore{} or \textallenmeets{} are considered \cite{ekvall2006learning, pardowitz2007incremental}.
They often allow for modeling and learning precedence task constraints in graphs or similar data structures, sometimes referred to as task precedence graphs.
Nicolescu and Matarić \cite{nicolescu2003natural} proposed a method to extract the longest common action sequence from several demonstrations of a task to identify unnecessary actions of a task.
These works are primarily based on a closed-world assumption where observed action sequences have to be adhered to for the robotic execution unless contradictions are observed.
While these approaches only account for precedences, Ye \etal{} \cite{ye2019robot} proposed a very similar approach that allows for parallel execution.
This is achieved under the assumption that actions in such a graph without any preceding action can be executed in parallel.
During task execution, a precedence graph is continuously updated with the remaining actions.
While this assumption can be valid in certain contexts (\ie scalable production lines), we argue that this is too relaxed in general, especially to model complex bimanual manipulation tasks.

Allen not only conceptualized temporal relations of two intervals in their work as discussed earlier, but also proposed algorithms to maintain temporal knowledge \cite{allen1983maintaining}.
However, both the works based on precedence graphs, and the one from Allen depend on perfect data to function properly.
Thus, in our previous work, we built on these ideas to learn \sttcs represented by Allen relations from bimanual human demonstrations \cite{dreher2022learning}.
Our approach can deal with imperfections of several kinds, such as not perfectly synchronized actions or errors in the demonstrations.
In contrast to our previous work, and the works discussed thus far, we want to additionally learn \ssttcs from human demonstrations, while still preserving the symbolic expressiveness of our previous work.

Cheng \etal{} \cite{cheng2021new} presented an algebraic foundation of Allen's interval algebra.
They observed that Allen relations are \emph{distinct} (no given pair of intervals can be part of more than one Allen relation), \emph{exhaustive} (any pair of intervals can be described by one Allen relation) and \emph{qualitative} (no subsymbolic or quantitative data of the intervals is preserved) \cite{cheng2021new}.
They proposed an algebraic system to extend Allen relations with many operations and so-called \emph{null intervals} to allow for the quantitative and complex description of temporal facts.
Instead of focusing on intervals, we look into the starts and ends (\ie time \emph{points}) of actions to model quantitative constraints between them, while still allowing to reason on a symbolic level using Allen's well-established interval algebra.

Another theoretical extension to Allen's interval algebra was proposed by Schockaert \etal{} \cite{schockaert2008fuzzifying} with the attempt to combine Allen's interval algebra with fuzzy set theory.
Instead, we estimate the degree of membership of time point differences to time point relations from observations and use fuzzy logic to model to which degree Allen relations hold given the observations.

Ramirez-Amaro \etal{} \cite{ramirez-amaro2013enhancing} implicitly learned spatio-temporal features from video data to improve an action recognition system.
This was achieved by utilizing an unsupervised neural network architecture based on Independent Subspace Analysis.
In contrast, we try to learn temporal constraints between actions of a task to gather and maintain temporal knowledge of a manipulation task.

\subsection{Planning Under Temporal Constraints}\label{ss:rw-temp-planning}

Many works on temporal planning require detailed data about mean action lengths and temporal variances, as well as temporal differences between starts and ends of actions.
Peller \etal{} \cite{peller2018temporal} proposed a planning system based on Temporal Fast Downward that can apply temporal stress on certain actions to meet task execution deadlines at the cost of higher task failure risk.
With the knowledge of which actions are safe to stress (indicated by a high temporal variance across several demonstrations), the risk of failure could be reduced if taken into consideration.
Fusaro \etal{} \cite{fusaro2021integrated} addressed the problem of planning and allocating tasks in the context of human-robot-collaboration.
They used Behavior Trees to model a task, encoding temporal and logic constraints of it.
This is achieved through Behavior Tree nodes that model sequential or parallel execution of actions.
The work focused on planning and allocating agents to tasks rather than learning these constraints.
Kantaros \etal{} \cite{kantaros2022perceptionbased} proposed a perception-based temporal planning system in the context of multi-robot navigation on semantic maps that used LTL for task specification. 
From this user-defined specification, an optimal control problem was formulated that allowed for online replanning while accounting for different kinds of uncertainties.

\subsection{Other Approaches}\label{ss:rw-causal-constraints}

Puranic \etal{} \cite{puranic2021learning} proposed a method based on Signal Temporal Logic (STL) to assess the quality of demonstrations, as well as to infer rewards for a Reinforcement Learning problem in the context of Programming by Demonstration and Inverse Reinforcement Learning.
STL is an extension to LTL that allows for continuous-time variables and real-valued predicates (signals) instead of discrete time variables and Boolean predicates as is the case in LTL.
Their basic idea is that demonstrations can be evaluated from user-defined temporal constraints via an STL specification.
Our work exactly aims at learning such temporal constraints between actions of a task from human demonstration.
Diehl and Ramirez-Amaro \cite{diehl2023causalbased} proposed a method to learn a causal Bayesian Network model from simulation to predict and avoid failures in task executions.
To obtain the causal model, two causal structure learning methods were compared.
They reported that, depending on the model, $\SI{200000}{}$ to $\SI{600000}{}$ samples were required for the models to converge for a stacking task with 3 cubes.
Instead, we learn temporal constraints that can be seen as a heuristic for causality when not enough knowledge is available (yet) from the provided demonstrations.
Especially in the context of learning from \emph{human} demonstration, so many samples are not feasible.

\section{Fundamentals}\label{s:fundamentals}

We provide definitions that are necessary to better understand our approach. This covers the definition of time point relations, time interval relations, and fuzzy logic. 

\subsection{Time Point Relations}\label{ss:time-point-rels}

A \emph{time point} $t\!\in\!\mathbb{R}$ is measured in seconds and a \emph{temporal difference} or \emph{duration} $d$ is the difference between two time points $d\!=\!t_1 - t_2$.
We will use the notation $\pointrel(t_1, t_2)\!\Leftrightarrow\!\relcondition$ to express that two time points $t_1, t_2\!\in\!\mathbb{R}$ are part of the \emph{time point relation} $\pointrel$ iff the condition $\relcondition$ holds.
Possible relations are $\pointbefore$, $\pointequals$, and $\pointafter$.
The definitions and examples for these relations can be seen in Table~\ref{tab:time-point-relations}.

\newcommand{\mr}[1]{\multirow{2}{*}{#1}}
\newcommand{\inv}[1]{\textcolor{black!66}{#1}}
\definecolor{xred}{RGB}{31,119,180}
\definecolor{ygreen}{RGB}{255,127,14}
\newcommand{\x}{\textcolor{xred}{x}}
\newcommand{\y}{\textcolor{ygreen}{y}}

\newcommand{\plotpoints}[2]{%
\begin{tikzpicture}[scale=0.5]
    \draw[white] (-0.5, 0.5) -- (4, 0.5);
    \draw[dashed, line width=0.1mm, gray] (0, 0.55) -- (4, 0.55);
    \draw[dashed, line width=0.1mm, gray] (0, 0.1) -- (4, 0.1);
    \filldraw[xred!15] (#1, 0.55) circle (1mm);
    \draw[line width=0.1mm, xred] (#1, 0.55) circle (1mm);
    \filldraw[xred] (#1, 0.55) circle (0.25mm);
    \node at (#1+0.25, 0.75) {\tiny $\x$};
    \filldraw[ygreen!15] (#2, 0.1) circle (1mm);
    \draw[line width=0.1mm, ygreen] (#2, 0.1) circle (1mm);
    \filldraw[ygreen] (#2, 0.1) circle (0.25mm);
    \node at (#2+0.25, 0.3) {\tiny $\y$};
\end{tikzpicture}%
}

\newcommand{\plotinterval}[4]{%
\begin{tikzpicture}[scale=0.5]
    \draw[white] (-0.5, 0.5) -- (4, 0.5);
    \draw[dashed, line width=0.1mm, gray] (0, 0.55) -- (4, 0.55);
    \draw[dashed, line width=0.1mm, gray] (0, 0.1) -- (4, 0.1);
    \draw[line width=1mm, xred!15] (#1, 0.55) -- (#2, 0.55);
    \draw[line width=0.25mm, xred] (#1, 0.55) -- (#2, 0.55);
    \draw[line width=0.1mm, xred] (#1, 0.65) -- (#1, 0.45);
    \draw[line width=0.1mm, xred] (#2, 0.65) -- (#2, 0.45);
    \node at (#1+0.25, 0.75) {\tiny $\x$};
    \draw[line width=1mm, ygreen!15] (#3, 0.1) -- (#4, 0.1);
    \draw[line width=0.25mm, ygreen] (#3, 0.1) -- (#4, 0.1);
    \draw[line width=0.1mm, ygreen] (#3, 0.2) -- (#3, 0);
    \draw[line width=0.1mm, ygreen] (#4, 0.2) -- (#4, 0);
    \node at (#3+0.25, 0.3) {\tiny $\y$};
\end{tikzpicture}%
}

\setlength{\tabcolsep}{2pt}
\begin{table}[ht]
    \centering
    \caption{All 3 time point relations.}\label{tab:time-point-relations}
    \vspace*{-2mm}
    \begin{tabular}{rcll}
        \toprule%
        Time Point Relation                            &                          & ~~~\mr{Condition}                          & ~~\mr{Example}                \\
        \inv{\scriptsize Inverse Relation}             &                          &                                            &                               \\
        \midrule%
        $\pointbefore(\x, \y)$                         & \mr{$\Leftrightarrow$}   & \mr{$~~~\x < \y$}                          & \mr{\plotpoints{1}{3}}~~      \\
        \inv{\scriptsize $\pointafter(\y, \x)$}        &                          &                                            &                               \\
        \greymidrule%
        $\pointequals(\x, \y)$                         & \mr{$\Leftrightarrow$}   & \mr{$~~~\x = \y$}                          & \mr{\plotpoints{2}{2}}        \\
        \inv{\scriptsize $\pointequals(\y, \x)$}       &                          &                                            &                               \\
        \bottomrule%
    \end{tabular}
    \vspace*{-3mm}
\end{table}
\setlength{\tabcolsep}{6pt}

\subsection{Time Interval Relations}\label{ss:time-interval-rels}

A \emph{time interval} is a tuple $\interval\!=\![\intstart{\interval}, \intend{\interval}]$ with $\intstart{\interval}$ and $\intend{\interval}$ being time points and $\intstart{\interval}\!<\!\intend{\interval}$.
Here, $\intstart{\interval}$ is the start of the interval $\interval$, and $\intend{\interval}$ is its end.
In the following, we will use the notation $\allenrel(\intervala, \intervalb)\!\Leftrightarrow\!\relcondition$ as a short-hand to express that two time intervals $\intervala, \intervalb\!\in\!I$, with $I$ being the set of all time intervals, are part of the \emph{time interval relation} $\allenrel$ iff the condition $\relcondition$ holds.
All possible relations between two time intervals have been described by Allen \cite{allen1983maintaining}, also called Allen's interval algebra or simply Allen Relations.
Their formal definition with examples of these relations can be seen in Table~\ref{tab:allen-original}.
Note that we included additional conditions in gray that follow from the definition of an interval and are thus redundant, but will become important later.
Additionally, note that we can plug the definitions of temporal time point relations from Table~\ref{tab:time-point-relations} into the conditions of the respective Allen relations.
\Eg we can express the time interval relation \textallenstarts{} as such: $ \allenstarts(\x, \y)\!\Leftrightarrow\!\pointequals(\intstart{\x}, \intstart{\y})\!\wedge\!\pointbefore(\intend{\x}, \intend{\y}) $.

\newcommand{\cfo}[1]{\textcolor{lightgray}{#1}}
\newcommand{\gwedge}{\mathbin{\cfo{\wedge}}}

\setlength{\tabcolsep}{2pt}
\begin{table}[ht]
    \vspace*{-1mm}
    \centering
    \caption{All 13 time interval relations after Allen \cite{allen1983maintaining}.}\label{tab:allen-original}
    \vspace*{-2mm}
    \begin{tabular}{rcll}
\toprule%
~~~Allen Relation                             &                          & ~~~\mr{Condition}                                                                    & ~~\mr{Example}                       \\
\inv{\scriptsize Inverse Relation}             &                          &                                                                                      &                                      \\
\midrule%
$\allenbefore(\x, \y)$                         & \mr{$\Leftrightarrow$}   & $~~~\cfo{\intstart{x} < \intstart{y}} \gwedge \cfo{\intstart{x} < \intend{y}}$       & \mr{\plotinterval{0}{1.5}{2.5}{4}}~~ \\
\inv{\scriptsize $\allenafter(\y, \x)$}        &                          & $\gwedge~\intend{\x} < \intstart{\y} \gwedge \cfo{\intend{x} < \intend{y}}$          &                                      \\
\greymidrule%
$\allenmeets(\x, \y)$                          & \mr{$\Leftrightarrow$}   & $~~~\cfo{\intstart{x} < \intstart{y}} \gwedge \cfo{\intstart{x} < \intend{y}}$       & \mr{\plotinterval{0}{2}{2}{4}}       \\
\inv{\scriptsize $\allenmetby(\y, \x)$}        &                          & $\gwedge~\intend{\x} = \intstart{\y} \gwedge \cfo{\intend{x} < \intend{y}}$          &                                      \\
\greymidrule%
$\allenoverlaps(\x, \y)$                       & \mr{$\Leftrightarrow$}   & $~~~\intstart{\x} < \intstart{\y} \gwedge \cfo{\intstart{x} < \intend{y}}$           & \mr{\plotinterval{0}{2.5}{1.5}{4}}   \\
\inv{\scriptsize $\allenoverlappedby(\y, \x)$} &                          & $\wedge~\intend{\x} > \intstart{\y} \wedge \intend{\x} < \intend{\y}$                &                                      \\
\greymidrule%
$\allenstarts(\x, \y)$                         & \mr{$\Leftrightarrow$}   & $~~~\intstart{\x} = \intstart{\y} \gwedge \cfo{\intstart{x} < \intend{y}}$           & \mr{\plotinterval{0}{2}{0}{4}}       \\
\inv{\scriptsize $\allenstartedby(\y, \x)$}    &                          & $\gwedge~\cfo{\intend{x} > \intstart{y}} \wedge \intend{\x} < \intend{\y}$           &                                      \\
\greymidrule%
$\allenduring(\x, \y)$                         & \mr{$\Leftrightarrow$}   & $~~~\intstart{\x} > \intstart{\y} \gwedge \cfo{\intstart{x} < \intend{y}}$           & \mr{\plotinterval{1}{3}{0}{4}}       \\
\inv{\scriptsize $\allencontains(\y, \x)$}     &                          & $\gwedge~\cfo{\intend{x} > \intstart{y}} \wedge \intend{\x} < \intend{\y}$           &                                      \\
\greymidrule%
$\allenfinishes(\x, \y)$                       & \mr{$\Leftrightarrow$}   & $~~~\intstart{\x} > \intstart{\y} \gwedge \cfo{\intstart{x} < \intend{y}}$           & \mr{\plotinterval{2}{4}{0}{4}}       \\
\inv{\scriptsize $\allenfinishedby(\y, \x)$}   &                          & $\gwedge~\cfo{\intend{x} > \intstart{y}} \wedge \intend{\x} = \intend{\y}$           &                                      \\
\greymidrule%
$\allenequals(\x, \y)$                         & \mr{$\Leftrightarrow$}   & $~~~\intstart{\x} = \intstart{\y} \gwedge \cfo{\intstart{x} < \intend{y}}$           & \mr{\plotinterval{0}{4}{0}{4}}       \\
\inv{\scriptsize $\allenequals(\y, \x)$}       &                          & $\gwedge~\cfo{\intend{x} > \intstart{y}} \wedge \intend{\x} = \intend{\y}$           &                                      \\
\bottomrule%
    \end{tabular}
    \vspace*{-3mm}
\end{table}
\setlength{\tabcolsep}{6pt}

\subsection{Fuzzy Set Theory\removed{, Infinite-Valued \Lukasiewicz Logic,} and Fuzzy Logic}\label{ss:fuzzy-logic}

\removed{\emph{Fuzzy set theory}, originally proposed by Zadeh \cite{zadeh1965fuzzy}, is an extension to traditional set theory.
Usually, a binary membership function $\chi_A(x)\!\in\!\{\text{\emph{false}}, \text{\emph{true}}\}$ describes whether the element $x$ is a member of the set $A$ or not.
Fuzzy set theory acknowledges, that often such a binary membership is not expressive enough and extends traditional set theory by defining the membership function $\mu_{\fuzzy{A}}(x)\!\in\![0, 1]$ as a real-valued function on the unit interval instead \cite{pykacz2000lukasiewicz}.
The value of such a membership function $\mu_{\fuzzy{A}}(x)$ is referred to as \emph{degree of membership} of the element $x$ to the fuzzy set $\fuzzy{A}$.
As a consequence, a \emph{fuzzy relation} is a relation on fuzzy sets \cite{zadeh1965fuzzy}.
To better distinguish fuzzy sets and fuzzy relations from traditional sets and relations, the terms \emph{crisp set} or \emph{crisp relation} are often used for the latter.}

\removed{Another related body of work is that of \emph{infinite-valued logic}, largely driven by \Lukasiewicz \cite{lukasiewicz1970selected}.
This work aimed to generalize traditional two-valued (Boolean) logic to allow for predicates to assume an infinite amount of values instead of two.
According to Pykacz \cite{pykacz2000lukasiewicz}, Giles \cite{giles1976lukasiewicz} later found ``\emph{the relation between fuzzy sets and infinite-valued \Lukasiewicz logic is exactly the same as the relation between traditional crisp sets and the classical two-valued} [Boolean] \emph{logic}''.}

\removed{\emph{Fuzzy logic} \cite{zadeh2008there} extends Boolean logic very similarly in that predicates can assume values from the unit interval $[0, 1]$, also referred to as \emph{degree of truth} or \emph{vagueness}.
Zadeh motivates that ``[\dots] \emph{fuzzy logic may be viewed as an attempt at formalization/mechanization of} [\dots] \emph{remarkable human capabilities}'', one of which is ``[\dots] \emph{the capability to converse, reason and make rational decisions in an environment of imprecision, uncertainty, incompleteness of information, conflicting information, partiality of truth and partiality of possibility -- in short, in an environment of imperfect information} [\dots]'' \cite{zadeh2008there}.}

\added{\emph{Fuzzy set theory} is an extension of traditional set theory that relaxes the assumptions made on set membership functions~\cite{zadeh1965fuzzy}.
In traditional set theory, a membership function $\chi_A(x)\!\in\!\{\text{\emph{false}}, \text{\emph{true}}\}$ is binary and describes whether the element $x$ is a member of the set $A$ or not.
Fuzzy set theory relaxes this assumption to allow for real-valued membership functions $\mu_{\fuzzy{A}}(x)\!\in\![0, 1]$.
Its value is referred to as \emph{degree of membership} of the element $x$ to the fuzzy set $\fuzzy{A}$.
As a consequence, a \emph{fuzzy relation} is a relation on fuzzy sets \cite{zadeh1965fuzzy}.
To better distinguish fuzzy sets and fuzzy relations from traditional sets and relations, the terms \emph{crisp set} or \emph{crisp relation} are often used for the latter.
Similarly to set theory, Boolean logic can also be extended to allow for real-valued, or fuzzy, variables and logic operations on them \cite{zadeh2008there,lukasiewicz1970selected}.}

In this work, we build on these ideas to understand time point and interval relations of actions in manipulation tasks not as crisp relations as defined in Sections~\ref{ss:time-point-rels} and \ref{ss:time-interval-rels}, but instead as fuzzy relations from observations made from human demonstrations.
These observations are inherently uncertain and conflicting due to many sources of possible errors, such as inaccuracies of the action segmentation or even errors in the human demonstration.

\section{Problem Formulation}\label{s:problem-formulation}

To provide a formal description of the problem, we start with important definitions of a task, an action, and a demonstration (Section~\ref{ss:defs-task-action-demonstration}) as well as types of temporal constraints considered in this work (Section~\ref{ss:temporal-constraints}).
Finally, we conclude with the problem statement in Section~\ref{ss:problem-statement}.

\subsection{Definition of Task, Action, Demonstration}\label{ss:defs-task-action-demonstration}

A task consists of a set of actions $\{\act_1, \act_2, \ldots\}$, where an action $\alpha$ is a tuple $\act\!=\!(v, o)$, with $v$ being a verb, and $o$ an object.
Let $\mathcal{D}$ be a set of demonstrations of a bimanual manipulation task. A \emph{demonstration} $D\!\in\!\mathcal{D}$ is a tuple $D\!=\!(A_L, A_R)$ consisting of two action sequences $A_L$ for the left hand, and $A_R$ for the right hand respectively.
An \emph{action sequence} $A$ is an ordered collection of action observations $A\!=\!(\actobs_1, \actobs_2, \actobs_3, \ldots)$.
An \emph{action observation} $\actobs$ is a tuple $\actobs\!=\!(\act, \interval)$ with $\act$ being an action and $\interval$ an interval $\interval\!=\![\intstart{\interval}, \intend{\interval}]$ with start time point $\intstart{\interval}$ and end time point $\intend{\interval}$.
We use $\intstart{\actobs}\!=\!\intstart{\interval}$ and $\intend{\actobs}\!=\!\intend{\interval}$ to refer to the two semantic temporal keypoints of the action observation $\actobs\!=\!(\act, \interval)$, namely its corresponding interval's $\interval$ start and end.
Action observations $\actobs_1, \actobs_2, \actobs_3, \ldots$ in an action sequence $A\!=\!(\actobs_1, \actobs_2, \actobs_3, \ldots)$ have a strict order and cannot overlap, thus it follows that $\intstart{\actobs_1}\!<\!\intend{\actobs_1}\!\leq\!\intstart{\actobs_2}\!<\!\intend{\actobs_2}\!\leq\!\intstart{\actobs_3} \ldots$

We assume that one hand can only execute one action at a time, and one action can only be executed by one hand.
A bimanual symmetric action in a demonstration $D\!=\!(A_L, A_R)$ (\eg lifting a bowl with both hands simultaneously, cf. \cite{krebs2022bimanual}), is represented by two actions observations $\actobs_L\!=\!(\act, \interval_L)\!\in\!A_L$ and $\actobs_R\!=\!(\act, \interval_R)\!\in\!A_R$ with the same action $\act$ and $\interval_L\!\approx\!\interval_R$.

\subsection{Types of Temporal Constraints}\label{ss:temporal-constraints}

In this work, we differentiate between several types of constraints to model conditions that need to be fulfilled on a symbolic and subsymbolic level for successful task execution.

A \emph{symbolic temporal task constraint} (\sttc) $\constrs\!=\!\allenrel$ is modeled through a time interval relation $\allenrel$ between two actions $\acta$ and $\actb$ of the task, \ie one Allen relation.
The set of \sttcs $\Constrs\!=\!\{\constrs_1, \constrs_2, \constrs_3, \ldots\}$ of a task must be free of contradictions for it to be useful.
They form the symbolic output of our approach and can be used for (re-)planning or temporal reasoning as shown in our previous work \cite{dreher2022learning}.
 
A \emph{subsymbolic temporal task constraint} (\ssttc) $\constrss\!\sim\!\mathcal{N}(\mu, \sigma^2)$ describes a temporal difference constraint between two semantic action keypoints $t_1$ and $t_2$ as a random variable of observed temporal differences from several demonstrations following a Gaussian distribution $\mathcal{N}(\mu, \sigma^2)$, where $\mu$ is the mean temporal difference and $\sigma^2$ the variance between $t_1$ and $t_2$.
The set of \ssttcs $\Constrss$ is defined as $\Constrss\!=\!\{\constrss_1, \constrss_2, \constrss_3, \ldots\}$.
They form the subsymbolic output of our approach and can be used to synchronize two MPs to generate a bimanual action execution from two unimanual actions.

For each \sttc $\constrs\!=\!\allenrel\!\in\!\Constrs$, we expect to obtain up to four \ssttcs $\constrss_1\!\sim\!\mathcal{N}(\mu, \sigma^2), \ldots, \constrss_4\!\in\!\Constrss$ that quantify how to realize the symbolic constraint $\constrs$.
This is done by specifying, which temporal differences between semantic keypoints of two actions should be enforced, to which offset (via the Gaussian distribution's mean), and to which temporal stress (via the Gaussian distribution's variance).

\subsection{Problem Statement}\label{ss:problem-statement}

We address the problem of learning temporal task constraints from human demonstration.
Specifically, given a set of demonstrations $\mathcal{D}$ and observed temporal differences between action keypoints and temporal relations between actions, the task is to infer two types of temporal task constraint sets: 
\begin{enumerate*}[label=(\roman*)]
\item a set of \sttcs $\Constrs$ as a means to \eg replan in case of unforeseen problems, and 
\item a set of \ssttcs $\Constrss$, as a means to \eg synchronize the execution of two MPs for bimanual manipulation actions by a robot.
\end{enumerate*}

\section{Methods and Approach}\label{s:approach}

To tackle the problem formulated in Section~\ref{s:problem-formulation}, \ie the problem of inferring \sttcs and \ssttcs from bimanual human demonstrations, we start by describing, how data from multiple demonstrations is aggregated to a model of temporal differences between semantic action keypoints (Section~\ref{ss:aggregating-data}).
In Section~\ref{ss:inference-overview}, we give a conceptual overview of the three phases of inferring \sttcs and \ssttcs, which are discussed in detail thereafter.

\subsection{Aggregating Demonstration Data}\label{ss:aggregating-data}

\newcommand{\apkm}{APKM\xspace}
\newcommand{\apkms}{APKMs\xspace}

As described before, our work builds on the hypothesis that important temporal constraints on symbolic and subsymbolic levels can be inferred from temporal differences between semantic action keypoints.
Since we deal with multiple demonstrations that may display several ways on how a task can be executed (\eg different orders of execution of certain actions), our model needs to capture this.
Thus, a single Gaussian distribution is not sufficient.
Instead, we model the distribution of temporal differences between two semantic action keypoints of a pair of actions in the task across several demonstrations using a Gaussian mixture model (GMM) $\gmm$.
We do this for each combination of semantic action keypoints of an action pair, and for each pair of actions in a task:
\begin{align*}
    \begin{split}
    \gmmsab = (\gmmasbs, \gmmasbe, \gmmaebs, \gmmaebe)
    \end{split}
\end{align*}
Here, $\gmmsab$ is the Action Pair Keypoint Model (\apkm), a 4-tuple of GMMs for all possible combinations of temporal differences between semantic action keypoints (cf. Figure~\ref{fig:facefig}).
The first element, $\gmmasbs$ is a GMM created from the set of all observed temporal differences between the starts of observations of action $\acta$ and the starts of observations of $\actb$ referred to as $\gmmdataasbs$.
Similarly, $\gmmasbe$ is created from $\gmmdataasbe$, $\gmmaebs$ from $\gmmdataaebs$, and $\gmmaebe$ from $\gmmdataaebe$, with
\newcommand{\forallactionobservationsvar}{\mathcal{A}_{\actb}^{\acta}}
\newcommand{\forallactionobservations}{\mid (i_j, i_k) \in \forallactionobservationsvar}%
{\allowdisplaybreaks
\begin{align*}
    \begin{split}
    \gmmdataasbs = \{\intstart{\interval_j} - \intstart{\interval_k} \forallactionobservations \},
    \end{split}\\
    \begin{split}
    \gmmdataasbe = \{\intstart{\interval_j} - \intend{\interval_k} \forallactionobservations \},
    \end{split}\\
    \begin{split}
    \gmmdataaebs = \{\intend{\interval_j} - \intstart{\interval_k} \forallactionobservations \},
    \end{split}\\
    \begin{split}
    \gmmdataaebe = \{\intend{\interval_j} - \intend{\interval_k} \forallactionobservations \},
    \end{split}%
\end{align*}}%
\begin{align*}
    \begin{split}
    \text{and~} \forallactionobservationsvar = \{ (i_j, i_k) \mid & \forall D = (A_L, A_R) \in \mathcal{D}, \\%
                                                                   & \forall (\actobs_j, \actobs_k) \in (A_L \cup A_R)^2, \\%
                                                                   & \actobs_j = (\acta, i_j), \actobs_k = (\actb, i_k) \},
    \end{split}%
\end{align*}%
where $\actobs\!=\!(\act, i)$ is an observation of action $\act$, and $X^2$ is the Cartesian product $X\!\times\!X$ of a sequence $X$.

To find the optimal number of components $N$ for each GMM, we employ an elbow-method-like approach.
For each $N\!=\!1, 2, ..., 10$ we compute the GMM and calculate the corresponding Bayesian Information Criterion (BIC) which describes, how well the model explains the data.
We chose that GMM for which the BIC is minimized.
\added{Please note that we allow for more than $3$ components in the GMM in this work. 
This allows for modeling cases where two actions show one qualitative relation (\eg  \textallenbefore) but in two different quantitative manifestations (\eg once with \SI{2}{\second} and once with \SI{4}{\second} between the actions).}

\subsection{Inferring Temporal Task Constraints: Outline}\label{ss:inference-overview}

Inferring temporal constraints from a set of \apkm can be divided into three steps.

\subsubsection{Estimating Degrees of Memberships to Temporal Relations}

As a first step, we are concerned with the question: \emph{Given multiple demonstrations of a task, how likely are certain temporal interval relations between each pair of actions?}
This means that we are interested in the degree of membership of each given pair of actions to all fuzzy time interval relations as a result of this step by analyzing the corresponding \apkm.
This research question is addressed in Section~\ref{ss:estimating-degrees-of-membership}.

\subsubsection{Inferring Symbolic Temporal Task Constraints}

In the second step, we are concerned with the question: \emph{Given the most likely temporal interval relations between each pair of actions of the task at hand, what is the single most likely temporal relation for each pair of actions in the task that is consistent within the whole temporal task model?}
Answering this question means assigning exactly zero or one temporal relation to each pair of actions in the task that are most likely, yet not in contradiction.
It can be seen as a fuzzy assignment problem.
This research question is addressed in Section~\ref{ss:infer-s-tmp-constraints}. 

\subsubsection{Inferring Subsymbolic Temporal Task Constraints}

In the third step, and given the knowledge from the two previous steps, this step addresses the question: \emph{How can we make use of temporal task constraints to synchronize MPs of a bimanual task execution precisely?}
Given a consistent set of temporal task constraints, we can leverage this information to discard components of the Gaussian mixture models of a pair of actions.
We assume that there is only one relevant mode in each GMM for execution because otherwise, it would violate the property of Allen relations being distinct.
This research question is addressed in Section~\ref{ss:infer-ss-tmp-constraints}.

\subsection{Estimating Degrees of Memberships to Temporal Relations}\label{ss:estimating-degrees-of-membership}

Given an \apkm $\gmmsab$ for each pair of actions $(\acta, \actb)$ in a task, we are interested in how distributions of temporal keypoint differences can be used to infer fuzzy time interval relations $\fallenrel$ describing the degree the observations are part of a given time interval relation $\allenrel$.

Similarly to how crisp time interval relations can be defined through time point relations as shown in Table~\ref{tab:allen-original}, we define fuzzy time interval relations through fuzzy time point relations $\fpointrel$.
With $\lukweakconj$ being the \Lukasiewicz weak conjunction operator $x\!\lukweakconj\!y\!=\!\min(x, y)$ \cite{pykacz2000lukasiewicz}, we define the degree of membership of the action pair $(\alpha_1, \alpha_2)$ of a task to a fuzzy time interval relation $\fallenrel$ through the corresponding \apkm $\gmmsab$ as follows:
{\allowdisplaybreaks
\begin{align*}
    \begin{split}
        \fallenbefore(\gmmsab) =~~~& \fpointbefore(\gmmasbs) \lukweakconj \fpointbefore(\gmmasbe) \\
                      \lukweakconj & \fpointbefore(\gmmaebs) \lukweakconj \fpointbefore(\gmmaebe)
    \end{split}\\
    \begin{split}
        \fallenmeets(\gmmsab) =~~~& \fpointbefore(\gmmasbs) \lukweakconj \fpointbefore(\gmmasbe) \\
                     \lukweakconj & \fpointequals(\gmmaebs) \lukweakconj \fpointbefore(\gmmaebe)
    \end{split}\\
    \begin{split}
        \fallenoverlaps(\gmmsab) =~~~& \fpointbefore(\gmmasbs) \lukweakconj \fpointbefore(\gmmasbe) \\
                        \lukweakconj & \fpointafter(\gmmaebs)  \lukweakconj \fpointbefore(\gmmaebe)
    \end{split}\\
    \begin{split}
        \fallenstarts(\gmmsab) =~~~& \fpointequals(\gmmasbs) \lukweakconj \fpointbefore(\gmmasbe) \\
                      \lukweakconj & \fpointafter(\gmmaebs)  \lukweakconj \fpointbefore(\gmmaebe)
    \end{split}\\
    \begin{split}
        \fallenduring(\gmmsab) =~~~& \fpointafter(\gmmasbs) \lukweakconj \fpointbefore(\gmmasbe) \\
                      \lukweakconj & \fpointafter(\gmmaebs) \lukweakconj \fpointbefore(\gmmaebe)
    \end{split}\\
    \begin{split}
        \fallenfinishes(\gmmsab) =~~~& \fpointafter(\gmmasbs) \lukweakconj \fpointbefore(\gmmasbe) \\
                        \lukweakconj & \fpointafter(\gmmaebs) \lukweakconj \fpointequals(\gmmaebe)
    \end{split}\\
    \begin{split}
        \fallenequals(\gmmsab) =~~~& \fpointequals(\gmmasbs) \lukweakconj \fpointbefore(\gmmasbe) \\
                        \lukweakconj & \fpointafter(\gmmaebs) \lukweakconj \fpointequals(\gmmaebe) 
    \end{split}%
\end{align*}}%
In addition to the characteristic conditions from the crisp time interval relations in Table~\ref{tab:allen-original}, we include all conditions since they have an impact on the calculation when using the \Lukasiewicz weak conjunction instead of a Boolean conjunction.
Please note that the inverse Allen relations are defined accordingly, but left out for brevity.

We define fuzzy time point relations $\fpointrel$ used in the definition above as follows.
Given one GMM $\gmm$ from the \apkm $\gmmsab$, let $\gmmcomponents{\gmm}$ be the set of components of $\gmm$, $\gmmcomponent{\gmm}{i}\!\in\!\gmmcomponents{\gmm}$ the mixtures $i$-th component with $\gmmcomponent{\gmm}{i}\!\sim\!\mathcal{N}(\mu_i, \sigma_i^2)$, and $\mu_i$ and $\sigma_i$ the component's mean and standard deviation, respectively.
Further, let $\smallno$ be a small duration and $\eqmargin\!=\![-\smallno, \smallno]$ an equality margin within which two time points are considered to be simultaneous.
We then identify the components $\gmmcomponent{\gmm}{i}$ of the GMM $\gmm$ whose means $\mu_i$ are outside the equality margin $\eqmargin$ such that $\mu_i\!<\!-\smallno$ or $\smallno\!<\!\mu_i$.
Their set is denoted as $\gmmcomponentsne{\gmm}\!\subset\!\gmmcomponents{\gmm}$, representing observations when the two time points cannot be considered simultaneous.

\begin{figure}[ht!]
    \vspace*{-2mm}
    \centering
    \includegraphics[width=\linewidth]{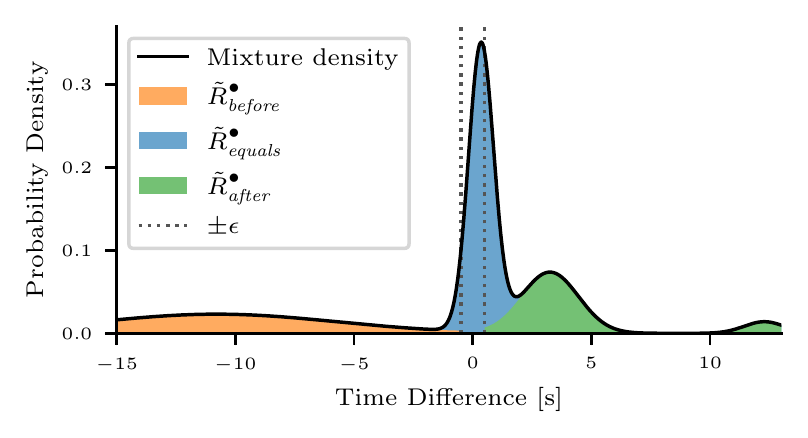}
    \vspace*{-10mm}
    \caption{To which degree a semantic temporal action keypoint $t_1$ is before, after or equal to another keypoint $t_2$ is decided by integrating the mixture density function $\gmmpdf{\gmm}$ of the GMM $\gmm$ that models the distribution of temporal differences between $t_1$ and $t_2$. 
    Red area: Degree of membership of $(t_1, t_2)$ in $\fpointbefore$. 
    Green area: Degree of membership of $(t_1, t_2)$ in $\fpointafter$. 
    Blue area: Degree of membership of $(t_1, t_2)$ in $\fpointequals$.}\label{fig:gmm-example}
    \vspace*{-4mm}
\end{figure}

Let $\gmmpdf{\gmm}$ be the mixture density function of the GMM $\gmm$: 
\begin{equation*}
    \gmmpdf{\gmm}(x) = \sum_{\gmmcomponent{\gmm}{i} \in \gmmcomponents{\gmm}} w_i \cdot{} p_i(x)
\end{equation*}
where $\gmmcomponent{\gmm}{i}$ is the $i$-th component of $\gmm$, $w_i$ its weight, $p_i$ it's probability density function.
We define $\gmmpdfne{\gmm}$ as mixture density function only considering components in $\gmmcomponentsne{\gmm}$:
\begin{equation*}
    \gmmpdfne{\gmm}(x) = \sum_{\gmmcomponent{\gmm}{i} \in \gmmcomponentsne{\gmm}} w_i \cdot{} p_i(x)
\end{equation*}
Using $\gmmpdfne{\gmm}$, we define the degrees of membership of the observed temporal differences between two semantic keypoints $t_1$ and $t_2$ modeled as GMM $\gmm$ to a fuzzy point relation $\fpointrel$:
\begin{align}
    \fpointbefore(\gmm) &= \int_{-\infty}^{-\smallno} \gmmpdfne{\gmm}(x) dx                   \label{eq:fpointrelbefore} \\
    \fpointafter(\gmm) &= \int_{\smallno}^{\infty} \gmmpdfne{\gmm}(x) dx                      \label{eq:fpointrelafter}  \\
    \fpointequals(\gmm) &= 1 - \fpointbefore(\gmm) - \fpointafter(\gmm)                       \label{eq:fpointrelquals}
\end{align}
Here, equation (\ref{eq:fpointrelbefore}) denotes the degree of membership to the $\fpointbefore$ relation.
This can be seen in shaded red in Figure~\ref{fig:gmm-example}.
Similarly, the degree of membership to $\fpointafter$ defined in equation (\ref{eq:fpointrelafter}) is shaded green,
the degree of membership to $\fpointequals$ in equation (\ref{eq:fpointrelquals}) is shaded blue, and the black curve shows the mixture density function $\gmmpdf{\gmm}$.

\subsection{Inferring Symbolic Temporal Task Constraints}\label{ss:infer-s-tmp-constraints}

In our previous work \cite{dreher2022learning}, we already presented an approach using a graph-theoretical approach by leveraging certain patterns occurring in bimanual manipulation tasks.
It takes a fully connected graph as input, where the nodes encode the actions of a task, and edges encode time interval relation probabilities that were directly assessed during observation.

In this work, we no longer assess the time interval relations during observation, but instead estimate the degree of membership of action observations to fuzzy time interval relations as shown in Section~\ref{ss:estimating-degrees-of-membership}.
However, we can still employ the same approach for inference, since the modalities have not changed.
Instead of encoding time interval relation probabilities, we now encode the degrees of membership to time interval relations.
The result is a set of \sttcs $\Constrs$ that are free of contradictions.

\subsection{Inferring Subsymbolic Temporal Task Constraints}\label{ss:infer-ss-tmp-constraints}

Given four GMMs of a pair of actions and the identified \sttc for that pair of actions, we now need to identify components of the GMM that are relevant for the task execution.
From the previous step (Section~\ref{ss:infer-s-tmp-constraints}) we obtained a contradiction-free assignment of \sttcs $\Constrs$.
They define qualitatively, how a subset of action pairs of the task are constrained through time interval relations $\allenrel$.
In this step, we are concerned with the question of how to quantitatively arrange the keypoints for execution to not only fulfill the relation qualitatively, but as closely to the demonstrations as possible.
For the scope of this work, the selection was done \replaced{by formulating a simple optimization problem}{naively by choosing the largest suitable component of relevant GMMs from the \apkm}.
\added{The relevant GMMs of an \apkm are those that correspond to the necessary conditions of \sttc (cf. conditions not grayed out in Table~\ref{tab:allen-original}).
Components are only suitable if their means have the expected sign given the necessary conditions.
For example to quantitatively arrange the keypoints of two actions with an \sttc of \textallenduring, only two temporal differences are necessary, namely those between both action's starts and both action's ends.
Thus, in the case of \textallenduring, only the largest components with a negative mean of the first GMM and the largest component with a positive mean of the fourth GMM from the \apkm are relevant.}

\section{Experiments and Evaluation}\label{s:eva}

In our previous work \cite{dreher2022learning}, we introduced two new benchmarks for learning \sttcs\footnote{\scriptsize{\href{https://bimanual-actions.humanoids.kit.edu/temporal_task_models}{\texttt{bimanual-actions.humanoids.kit.edu/temporal\_task\_models}}}}.
Hence, we compare how our proposed method performs in these benchmarks in Sections~\ref{ss:eva-syn} and \ref{ss:eva-real}. 
Additionally, we showcase the use of extracted \ssttcs in a qualitative evaluation in Section~\ref{ss:eva-subsymbolic-constraints}.

\subsection{Quantitative Evaluation of Inferred \sttcs}\label{ss:eva-syn}

\noindent\textbf{Experimental Setup:} For this evaluation, the synthetic dataset proposed in our previous work is used \cite{dreher2022learning}, which generated viable demonstrations using an exhaustive temporal planner given a user-defined set of actions and temporal constraints between them.
It consists of $216$ demonstrations of a disassembly scenario of an electric motor, with two different ways of accomplishing the task.
As in our previous work, we define a \emph{learning scenario} as a simulated learning process of a temporal task model.
Each learning scenario creates its own temporal task model. 
$100$ demonstrations out of $216$ are randomly assigned to each learning scenario. 
Demonstrations are added one by one, and the identified \sttcs are evaluated after each addition. 
This approach results in unique learning processes, with $100$ learning scenarios evaluated in total.
Comparing the ground truth \sttcs to those inferred by the temporal task models can yield $4$ different outcomes, as already described in \cite{dreher2022learning}:
\begin{enumerate*}[label=(\roman*)]
	\item \emph{True positive: A temporal constraint is identified, which is in the ground truth data.}
	\item \emph{False positive: A temporal constraint is hypothesized to be there, but it is not in the ground truth.}
	\item \emph{False negative: A temporal constraint is in the ground truth, but it was not identified.}
	\item \emph{True negative: No temporal constraints were identified between two actions, and there is no corresponding constraint in the ground truth.}
\end{enumerate*}
We report on the precision and recall scores and repeat the evaluation process for both the proposed approach, as well as the one in our previous work \cite{dreher2022learning}.
We favor false positives over false negatives.
Especially at the beginning, we expect the temporal task models to overestimate \sttcs because with few demonstrations, not many conclusions can be made.
But with more demonstrations, the temporal task model should be able to relax certain constraints as they can turn out to be coincidences.
\Eg if a robot always observes that the relation $\acta$ \textallenbefore{} $\actb$ holds, the robot should assume that this constraint must be adhered to.
If the robot later observes contradictory demonstrations (\ie $\actb$ \textallenbefore{} $\acta$), it may relax this constraint.
In any case, however, false negatives should be avoided, since these essentially mean that necessary constraints used for generating the dataset were ignored.

\begin{figure}[ht!]
    \vspace*{-3mm}
    \centering
    \includegraphics[width=\linewidth]{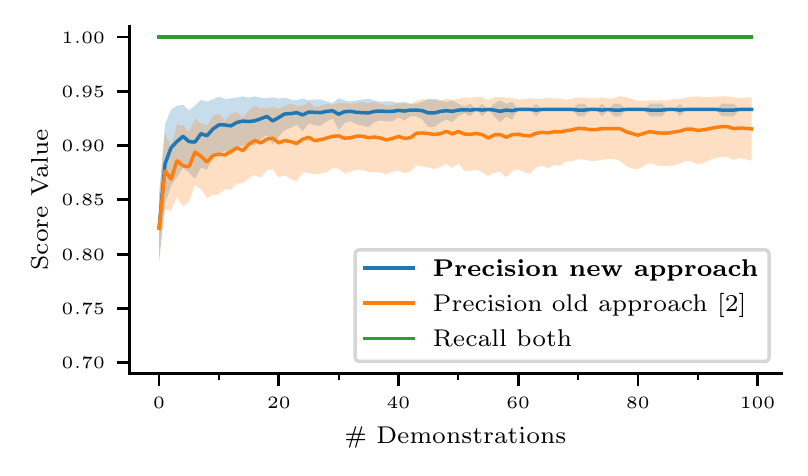}
    \vspace*{-8mm}
    \caption{Mean precision and recall with standard deviation of our new approach compared to our old approach \cite{dreher2022learning} over $100$ simulated learning scenarios. 
    Each learning scenario is an independent instantiation of a temporal task model that receives $100$ demonstrations of the synthetic dataset one by one in a unique order. 
    The temporal task model of each learning scenario is evaluated after adding a new demonstration.}\label{fig:eva-agiporbot-result}
    \vspace*{-2mm}
\end{figure}

\noindent\textbf{Results:} Figure~\ref{fig:eva-agiporbot-result} shows the mean precision and recall, as well as the standard deviation across all $100$ learning scenarios for both the proposed approach and our previous work~\cite{dreher2022learning}.
As can be seen, our new approach slightly outperforms the old approach.
Additionally, the variance of the new approach improves with more demonstrations, while the old approach shows a large variance regardless of the number of demonstrations.
This can be attributed to the more fine-granular method of obtaining \sttcs from temporal differences between semantic action keypoints instead of committing to \sttcs directly from the demonstration as in our old approach.
Both approaches maintain a recall of $1$ regardless of the number of demonstrations, supporting the design goal of not being too relaxed.

\subsection{Qualitative Evaluation of Inferred \sttcs}\label{ss:eva-real}

\noindent\textbf{Experimental Setup:} In this experiment, we use the re-labeled dataset of the KIT Bimanual Actions (Bimacs) Dataset~\cite{dreher2020learning} from our previous work~\cite{dreher2022learning} that contains $60$ demonstrations of $6$ subjects preparing a muesli in total.
The subjects were asked to prepare muesli given a banana\added{ (unpeeled, must be cut)}, a pack of cereals, a bottle of milk, and a bowl.
\removed{Additionally, a cutting board and a knife were on the table to cut the banana to put it as an ingredient into the muesli.}
In total, $4$ subtasks could be observed, namely \emph{cut banana, etc.}, \emph{pour banana, etc.}, \emph{pour cereals, etc.}, and \emph{pour milk, etc}.
Within these $60$ demonstrations, $8$ unique and unequally distributed variations of subtask sequences were observed (cf. \cite{dreher2022learning}, Table III).
As discussed in our previous work, ground truth data for this dataset is not available, thus this evaluation is performed qualitatively by testing, whether the model can identify, that the task consists of $4$ subtasks and that only \emph{cut banana etc.} \textallenbefore{} \emph{pour banana etc.} holds.
For comparison, this evaluation is performed for the synthetic dataset as well, expecting $5$ or $6$ subtasks in this case.

\begin{figure}[ht!]
    \vspace*{-3mm}
	\centering
	\includegraphics[width=\linewidth]{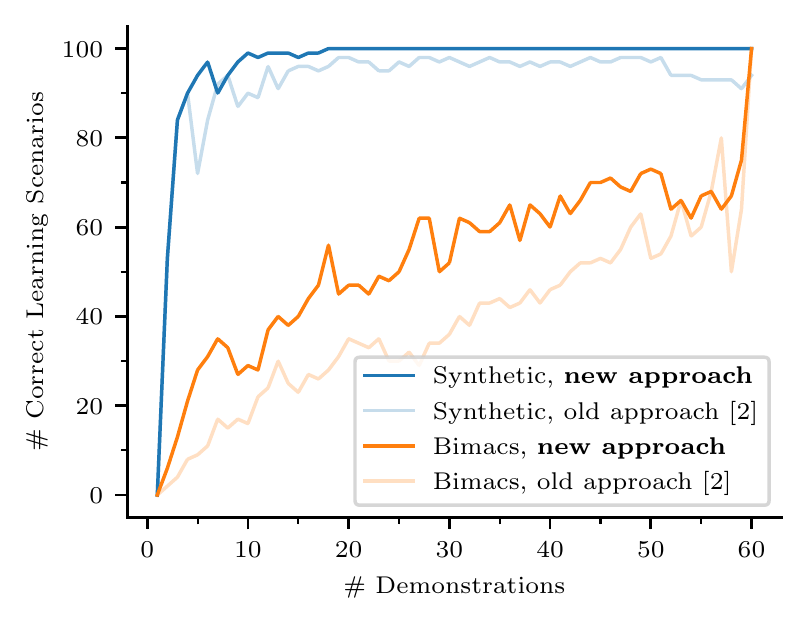}
    \vspace*{-8mm}
	\caption{Correct learning scenarios across $60$ demonstrations, comparing the new approach with that of our previous work \cite{dreher2022learning} for both the synthetic dataset (Synthetic), as well as the KIT Bimanual Actions Dataset (Bimacs).}\label{fig:eva-correct-combined}
    \vspace*{-2mm}
\end{figure}

\noindent\textbf{Results:} In Figure~\ref{fig:eva-correct-combined}, the number of correct learning scenarios depending on the number of demonstrations they received can be seen for the new approach compared to the old one for both datasets.
As can be seen, the new approach is generally able to perform better compared to our old approach.
The model performs better on the synthetic dataset because this dataset only consists of $2$ equally distributed modes of possible task executions compared to $8$ heavily unequally distributed modes in the Bimacs dataset.

\subsection{Showcase on Synchronizing Movement Primitives}\label{ss:eva-subsymbolic-constraints}

In this qualitative evaluation we showcase how the synchronization of MPs of two predefined actions of a pouring task from the KIT Bimanual Manipulation Dataset \cite{krebs2021kit} may take place using \ssttcs inferred using our approach.
This task involves two key actions executed bimanually, namely holding the cup while pouring the milk.
Prerequisites are an existing temporal task model learned from 20 random demonstrations of the pouring task, as well as existing MPs for these actions that were learned unimanually.
The MPs associated with the actions could come from an MP library, such as in our work~\cite{daab2024incremental}.
We are interested in synchronizing said two key actions.
\added{Our approach first identifies an \sttc of \textallenduring{} between the two key actions.
From this, the largest suitable component for each of the two relevant GMMs is selected.
These form two identified \ssttcs, namely that \emph{pour milk} must start \SI{0.5}{\second} before \emph{hold cup} starts, and that \emph{pour milk} must end \SI{0.5}{\second} after \emph{hold cup} ends.
The concrete temporal arrangement is done by choosing durations of the two key actions so that the two identified \ssttcs are satisfied.
To this end, we formulate an optimization problem}
\begin{align*}
    \begin{split}
    \added{min_{t_m,t_c} (\abs{t_m\!-\!t_c}\!-\!(\Delta^-\!+\!\Delta^+))\!+\!(t_m\!-\!t_M)\!+\!(t_c\!-\!t_C),}
    \end{split}
\end{align*}
\added{where $t_m$ and $t_c$ are the duration of the \emph{pour milk} and \emph{hold cup} MPs, respectively, and $t_M$ and $t_C$ the mean observed action duration of \emph{pour milk} and \emph{hold cup}.
Additionally, $\Delta^-$ is the temporal difference between the two key action's starts and $\Delta^+$ the temporal difference between the key action's ends (\ie the identified \ssttcs of \SI{0.5}{\second} each).}
The video attachment\footnote{\scriptsize{\href{https://youtu.be/ilIKP4DSF-E}{\texttt{youtu.be/ilIKP4DSF-E}}}} shows how pre-trained unimanual MPs were synchronized \added{as described} to realize a bimanual pouring action using the \ssttcs inferred using our approach.

\section{Conclusion and Future Work}\label{s:conclusion}

In this work, we presented an approach to learning symbolic and subsymbolic temporal task constraints between actions of a task from bimanual human demonstrations.
The foundation of such a temporal task model is a set of distributions of temporal differences between semantic action keypoints.
Specifically, we propose to employ Gaussian mixture models of temporal differences between action's starts and ends to capture the complex temporal nexuses of actions in bimanual human manipulation tasks.
A novel approach based on fuzzy logic was presented to define fuzzy Allen relations from fuzzy time point relations.

In future works, we want to extensively evaluate and validate our proposed approach in real robot experiments and study, how the now available \ssttcs can be utilized to their full potential to synchronize Via-point Movement Primitives~\cite{zhou2019learning} or facilitate their sequencing and blending~\cite{jaquier2022learning}.
Eventually, we envision holistic spatio-temporal task models that also capture spatial constraints both on a symbolic and subsymbolic level very similar to the temporal constraints in this work\removed{, while still incorporating this work's findings}.

\section*{Acknowledgment}

We would like to thank André Meixner for the valuable discussions, helpful feedback, and support.

\bibliographystyle{IEEEtran}
\bibliography{2023_combined_temporal_models}

\end{document}

%% file: h2t_def.tex
\usepackage[per-mode=fraction]{siunitx}
\usepackage{comment}
\usepackage[normalem]{ulem}
\usepackage{xspace}
\usepackage{xcolor}

\DeclareSIUnit\px{px}
\DeclareSIUnit\fps{fps}

\definecolor{OliveGreen}{RGB}{0,200,25}
\newcommand{\red}[1]{\textcolor{red}{#1}}

\newcommand{\darkgreen}[1]{\textcolor{OliveGreen}{#1}}

\newcommand{\ie}{i.\,e.,\xspace}
\newcommand{\eg}{e.\,g.,\xspace}

\newcommand{\Eg}{E.\,g.,\xspace}
\newcommand{\etal}{et\,al.\xspace}

\newif\iffinal

\newcommand{\enablefinalversion}{\finaltrue}

\newcommand{\added}[1]{%
	\iffinal%
	#1%
	\else%
	\darkgreen{#1}%
	\fi%
}
\newcommand{\replaced}[2]{%
	\iffinal%
	#2%
	\else%
	\red{\ifmmode\text{\sout{\ensuremath{#1}}}\else\sout{#1}\fi}\darkgreen{#2}%
	\fi%
}
\newcommand{\removed}[1]{%
	\iffinal%
	\else%
	\red{\ifmmode\text{\sout{\ensuremath{#1}}}\else\sout{#1}\fi}%
	\fi%
}